\documentclass[
]{ceurart}

\usepackage[frozencache=true,cachedir=minted-cache]{minted}
\usepackage{subcaption}
\setminted{breaklines=true}

\begin{document}
% !TeX TXS-program:compile = txs:///pdflatex/[--shell-escape]
%%
%% Rights management information.
%% CC-BY is default license.
\copyrightyear{2025}
\copyrightclause{Copyright for this paper by its authors.
  Use permitted under Creative Commons License Attribution 4.0
  International (CC BY 4.0).}

%%
%% This command is for the conference information
\conference{De-Factify - Fourth Workshop on Multimodal Fact-Checking and Hate Speech Detection}

%%
%% The "title" command
\title{AI Generated Text Detection Using Instruction Fine-tuned Large Language and Transformer-Based Models}

%%
%% The "author" command and its associated commands are used to define
%% the authors and their affiliations.
\author[1]{Chinnappa Guggilla}[%
email=cguggilla@deloitte.com
]
\address[1]{Deloitte \& Touche Assurance and Enterprise Risk Services India Private Limited, India}

\author[2]{Budhaditya Roy}[%
email=buroy@deloitte.com
]
\address[2]{Deloitte \& Touche LLP, USA}

\author[1]{Trupti Ramdas Chavan}[%
email=trchavan@deloitte.com
]
% \address[1]{Deloitte \& Touche Assurance and Enterprise Risk Services India Private Limited, India}

\author[2]{Abdul Rahman}[%
email=abdulrahman@deloitte.com
]
% \address[2]{Deloitte \& Touche LLP, USA, USA}

\author[2]{Edward Bowen}[%
email=edbowen@deloitte.com
]
% \address[2]{Deloitte \& Touche LLP, USA, USA}

% \maketitle
%%
%% The abstract is a short summary of the work to be presented in the
%% article.
\begin{abstract}
Large Language Models (LLMs) possess an extraordinary capability to produce text that is not only coherent and contextually relevant but also strikingly similar to human writing. They adapt to various styles and genres, producing content that is both grammatically correct and semantically meaningful. Recently, LLMs have been misused to create highly realistic phishing emails, spread fake news, generate code to automate cyber crime, and write fraudulent scientific articles. Additionally, in many real-world applications, the generated content including style and topic and the generator model are not known beforehand. The increasing prevalence and sophistication of artificial intelligence (AI)-generated texts have made their detection progressively more challenging. Various attempts have been made to distinguish machine-generated text from human-authored content using linguistic, statistical, machine learning, and ensemble-based approaches. This work focuses on two primary objectives\: Task-A, which involves distinguishing human-written text from machine-generated text, and Task-B, which attempts to identify the specific LLM model responsible for the generation. Both of these tasks are based on fine tuning of Generative Pre-trained Transformer (GPT\_4o-mini), Large Language Model Meta AI (LLaMA) 3 8B, and Bidirectional Encoder Representations from Transformers (BERT). The fine-tuned version of GPT\_4o-mini and the BERT model has achieved accuracies of 0.9547 for Task-A and 0.4698 for Task-B. 
% The highest accuracies in our experiments were recorded by the GPT\_4o-mini model for Task-A and the BERT model for Task-B.
\end{abstract}

%%
%% Keywords. The author(s) should pick words that accurately describe
%% the work being presented. Separate the keywords with commas.

%%
%% This command processes the author and affiliation and title
%% information and builds the first part of the formatted document.
\maketitle

\section{Introduction} \label{intro}
% The remarkable advancement in natural language processing, demonstrated by the emergence of large language models (LLMs) like Open AI's GPT-3 and GPT-4, Meta's LLaMA-2 and LLaMA-3, Google's Gemini and Anthropic's Claude has presented both opportunities and challenges.
The advancement in natural language processing, demonstrated by the emergence of large language models (LLMs) has presented both opportunities and challenges. 
% These LLMs, known for their capacity to generate highly realistic and fluent text, support a wide range of applications, including code-assist, conversational systems, question answering, machine translation and news article generations. While 
These models provide numerous beneficial applications; however, they also pose potential risks such as the widespread creation of synthetic disinformation, spam, and phishing content
 \citep{solaiman2019releasestrategiessocialimpacts} and fake news generation \citep{uchendu2021turingbenchbenchmarkenvironmentturing}.
To address these LLM challenges, a significant amount of work is being done to distinguish human-authored text from machine-generated text. Due to the ubiquitous nature of these neural language models, distinguishing machine-generated texts from human-written ones is no longer sufficient, and it is also necessary to identify which specific neural text generator authored a piece of text \citep{uchendu2021turingbenchbenchmarkenvironmentturing}.
To develop accurate detectors of machine-generated texts, sufficient data is essential. The authors \citep{uchendu2021turingbenchbenchmarkenvironmentturing} created the first benchmark for Authorship Attribution in the form of the Turing Test, incorporating both human and neural language models. 

There are several proprietary tools such as Giant Language Model Test Room (GLTR) \cite{aidet_tools}, GPT-2 output detector, and GPTZero. However, these tools act as a black box and there is very little understanding about their working principal. The human and AI text classification literature can be broadly categorized into supervised, zero shot, retrieval-based, watermarking, and feature-based detection methods \cite{abdali2024decoding, wu2024surveyllmgeneratedtextdetection}. In the supervised methods, the pre-trained models are fine tuned for classification while zero-shot detection methods apply the pre-trained model in zero-shot settings. The retrieval-based algorithms uses similarity between given text and LLM generations. The watermarking approaches utilize the model signature in the generated text, while feature-based methods consider various features for classification.
In the geneRative AI Detection viA Rewriting method (Raidar) \cite{mao2024raidargenerativeaidetection}, it is found that while using LLMs for the rewriting task, there is a high possibility of modification of human-generated text.
% it is found that while using LLMs for the rewriting task, the modification of human-generated text is highly possible than AI text. 
Also, to identify AI-generated text, LLMs were prompted to rewrite and the editing distance was calculated between the outputs. Harika A. et al. \cite{abburi2023simpleefficientensembleapproach}, used ensemble of five pre-trained transformer based models to generate the robust features and various ML algorithms such as logistic regression, gaussian naive bayes, random forest, and support vector machine were applied for text classification.
% This method is used as baseline for the evaluation of the proposed method in the testing set of Task A and B.

% To better understand which types of AI-generated content are easier or harder to detect is essential, our work aims to leverage dataset related to various families of LLMs, including Encoder models (e.g., BERT, DeBERTa), Open-Source models (e.g., Llama 3.1, Yi-2), Closed-Source models (e.g., Claude 3.5 Sonnet, GPT 4.0/mini), SLMs (e.g., Phi-3.5), Mixture of Experts (MoEs) (e.g., Mixtral), and State Space Models (SSM) (e.g., Falcon-Mamba). 

In this work, two sub-tasks, namely, Task-A and Task-B are focused. The goal of Task-A is to determine whether a given text document is generated by AI or created by a human. Task-B focuses on the identification of which specific LLM such as DeBERTa, Falcon, GPT 4.0 \citep{openai2023gpt4}, Mamba, Phi-3.5 has generated a given text. The Open AI's GPT-4o-mini \citep{openai2024gpt4omini}, LLaMA-3 8B pre-trained LLMs, and transformer-based BERT \citep{kenton2019bert} models are fine-tuned on the training datasets for Task-A and Task-B with simple prompts. The performance of the proposed approach is evaluated using a given validation and test dataset.

The organization of the paper is as follows. Section \ref{dataset} describes the dataset used in this work. The simple prompt-based instruction tuning methodology, and experimental results are presented in Sections \ref{prop_approach} and \ref{exp_results}, respectively. This work is concluded in Section \ref{conclusion}

\section{Dataset} \label{dataset}
This work uses the dataset provided by the Defactify workshop \citep{roy-2025-defactify-overview-text, roy-2025-defactify-dataset-text}. 
% The training dataset is publicly available on Huggingface \cite{roy-2025-defactify-dataset-text}. 
The validation and testing data sets are provided by the workshop organizers. The details of the dataset distribution are mentioned in Table \ref{tab:data}. 
% The data is comprised of human-written stories about various domains (medical, election, etc.) and their corresponding AI text generated by various LLMs. 
\subsection{Data Analysis}
\begin{figure}
\centering
\includegraphics[width=0.7\textwidth]{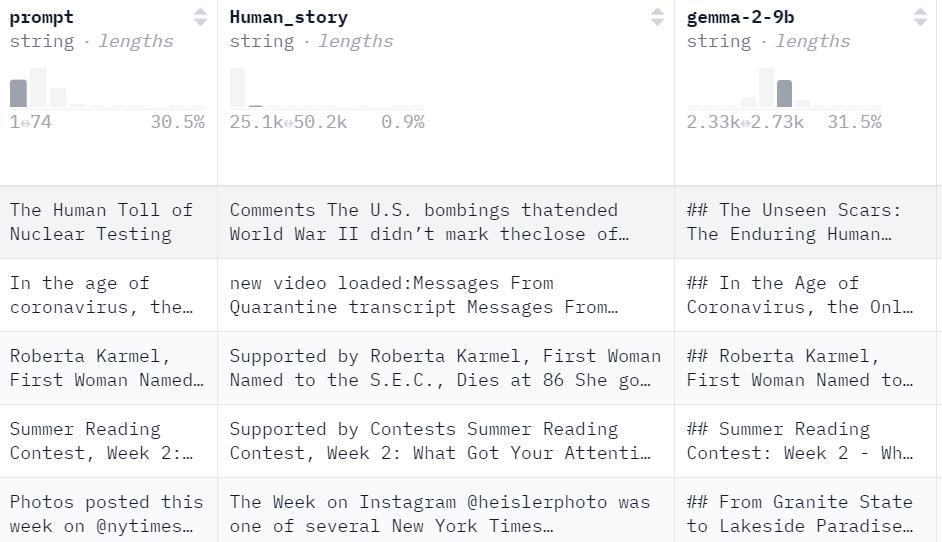} % Reduce the figure size so that it is slightly narrower than the column.
\caption{An illustration of the sample format includes an example prompt, a human-generated story, and text generated by the gemma-2-9b model from the text detection training dataset. \citep{roy-2025-defactify-dataset-text}} %The training data set sample format with example prompt, human generated story and gemma-2-9m model generated text.}
\label{fig1}
\end{figure}

The dataset includes human-authored stories from a variety of domains, including medical and election topics, as well as corresponding AI-generated text produced by several LLMs. The diverse set of LLMs used in the dataset creation process includes gemma-2-9b, GPT 4.0, llama-8b, mistral-7b, qwen2-72b and Yi-large. These models belong to LLMs families such as encoder, open-source, closed-source, small language models, mixture of experts and state space models. The snapshot of the dataset sample consisting of example prompt, human story and gemma-2-9b model generated text is shown in Figure \ref{fig1}. Figure \ref{datafig1} illustrates the lexical diversity, while Figure \ref{datafig2} shows the average length of sentences from the training dataset. The majority of the human and LLM-generated text samples in the training dataset have a sequence length of 0 to 100. Human-written samples exhibit a wider lexical diversity, ranging from 0.2 to 0.8, whereas LLM-generated model samples fall within the range of 0.4 to 0.6.
% Further the given train-set is preprocessed to get 7,321 'human' and 43,926 'machine-generated' samples to build binary and multi-class classifiers.

% \begin{figure}
%   \centering
%   \includegraphics[width=\linewidth]{sample-franklin}
%   \caption{1907 Franklin Model D roadster. Photograph by Harris \&
%     Ewing, Inc. [Public domain], via Wikimedia
%     Commons. (\url{https://goo.gl/VLCRBB}).}
% \end{figure}

The labels in train and validation set are equally distributed as shown in the Table \ref{tab:data}. The dataset consists of seven labels that includes human-written and six different LLMs. 
% provided in training and validation set are distributed and includes labels such as human-written and six differnt LLMs.
\begin{table}
  \caption{Distribution of text detection dataset \citep{roy-2025-defactify-dataset-text} for Task-A and Task-B}
  \label{tab:data}
  \begin{tabular}{cccc}
    \toprule
    Data set & Human & Machine  & Total \\
    \midrule
        Train & 7,321 & 43,926 & 51,147 \\
        Validation & 1,569 & 9,414 & 10,983\\
        Test & - & - & 10,963\\
 \bottomrule
  \end{tabular}
\end{table}

\section{Proposed Approach} \label{prop_approach}
In this section, we describe the proposed methodology for modeling Task-A and Task-B. Task-A is approached as a binary classification problem, where the goal is to determine whether a given input text is human-written or machine-generated. Task-B, on the other hand, aims to identify which LLM generated the input text and is treated as a multi-class classification problem. For Task-A, we fine-tuned OpenAI's GPT\_4o-mini and Google's BERT, whereas for Task-B, we fine-tuned BERT and META's LLAMA-3 8B. The fine-tuning approaches based on simple prompts are explained in the following subsections.

% \subsection{Modeling Task-A and Task-B as Binary Classification and Multi Classification Problems}
% We describe the prompting instruction, training dataset format modeling method and parameters that are being used during fine-tuning and inference stages for the Task-A.
\subsection{Instruction training data set for Task-A and Task-B.}
To fine tune GPT\_4o-mini and LLaMA-3 8B models, the instruction tuning dataset is prepared based on inclusion of the respective prompt instructions mentioned in Table \ref{prompt_tab} and the prompts given in the dataset are excluded. Due to the downtime of the Open AI endpoint, GPT\_4o-mini model was not fine-tuned for Task-B in our environment. For fine-tuning of pre-trained BERT model, the training dataset is prepared using input pair of text and labels.

\begin{table}
  \caption{Prompts used in the instruction tuned dataset for Task-A and Task-B}
  \label{prompt_tab}
\begin{tabular}{ccl}
\toprule
Task & Models                      & \multicolumn{1}{c}{Prompt}                                                                                                                                                                                                                                                                                                                                                                      \\
\midrule
A    & \begin{tabular}[c]{@{}c@{}}GPT\_4o-mini \\ and LLaMA-3 8B\end{tabular} & \textit{\begin{tabular}[c]{@{}l@{}}Given a human written text or machine generated text, \\ classify whether   the given text is written by 'human' or 'machine'.\end{tabular}}                                                                                                                                                                                                                 \\
\\

B    & LLaMA-3 8B                  & \textit{\begin{tabular}[c]{@{}l@{}}Given a human written text or machine generated text as input, \\ classify   the given input text into one of these 7 labels. \\ These labels are:  gemma-2-9b, GPT 4.0, Human\_story, llama-8b, mistral-7b, \\ qwen2-72b and Yi-large. These labels are different in the way \\ how the text is written, syntax and lexical diversity.\end{tabular}} \\
 \bottomrule
  \end{tabular}
\end{table}

\subsection{Fine-tuning of LLMs and BERT models}%GPT\_4o-mini Model for Task-A.}
\begin{figure*}[t]
\centering
\begin{subfigure}{0.45\linewidth}
    \includegraphics[width=0.9\linewidth]{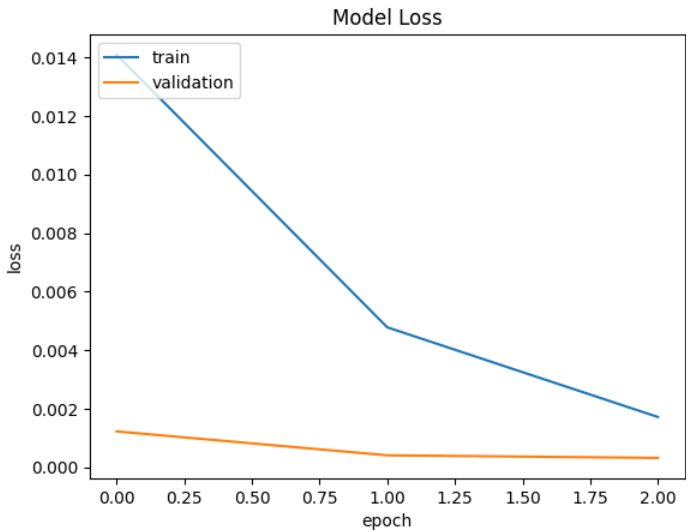}
    \caption{Loss curves for Task-A}
    \label{fig2}
\end{subfigure}
% \hfill
\begin{subfigure}{0.45\linewidth}
    \includegraphics[width=0.9\linewidth]{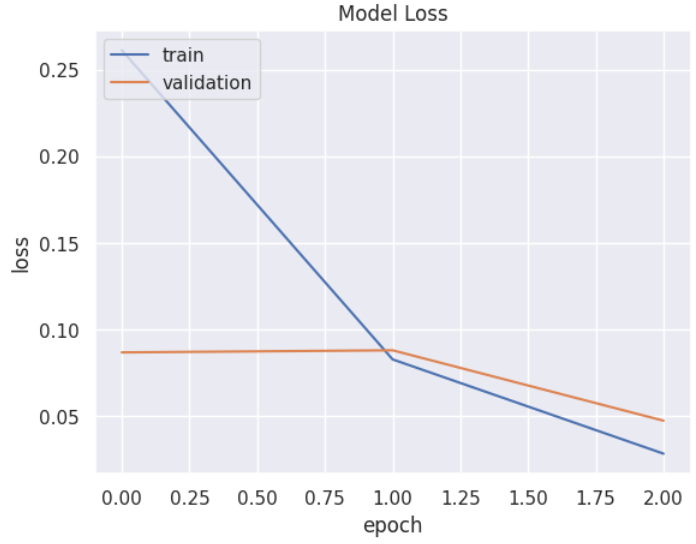}
    \caption{Loss curves for Task-B}
    \label{fig3}
\end{subfigure}
\caption{Train and validation losses of fine-tuned BERT model for Task-A and Task-B}
\label{train_val_loss}
\end{figure*}

The prompts shown in the Table \ref{prompt_tab} are appended to the input text along with the label names before being fed into the GPT\_4o and LLaMA-3 8B models for fine-tuning. After the models are fine-tuned, the same prompting instructions are utilized during inference on the test dataset. The final results shown in the Table \ref{table5} are obtained by combining GPT\_4O-mini Task-A results with BERT model Task-B results.
% For fine-tuning BERT pre-trained model, no prompt instruction is required and the training dataset is prepared using input pair of text and label.
The details of fine tuning of LLM models for Task-A and B are provided in Table \ref{fine_tune}
The training and validation losses for Tasks-A and B are plotted in Figure \ref{fig2} and Figure \ref{fig3}, respectively. Both train and validation losses are seen to decrease gradually until 3 epochs.

\begin{table}
  \caption{Experimental settings related to fine-tuning of GPT\_4o-mini, LLaMA-3 8B and BERT models for Task-A and B}
  \label{fine_tune}
\begin{tabular}{cclc}
\toprule
Model        & Task    & \multicolumn{1}{c}{Parameter Settings}                                                                                                                                                                                                                                                                                                                                                                                                                                                                        & Fine tuning Duration \\
\midrule
GPT\_4o-mini & A       & \begin{tabular}[c]{@{}l@{}}Azure Open AI model: GPT\_4o-mini \\      Model Version: gpt-4o-mini 2024-08-01-preview\\      Batch Size: 2 to 4;   Epochs: 1 to 2\end{tabular}                                                                                                                                                                                                                                                                                                              & 24 hours             \\
\\
LLaMA-3 8B   & A and B & \begin{tabular}[c]{@{}l@{}}GPU config: NVIDIA A100-SXM4-80GB \\ 
Model Version: llama-3-8b-bnb-4bit from Huggingface\\ Model parameter Optimization: Used Unsloth \cite{unsloth} and \\ its   FastLanguageModel library with 4-bit LoRA \cite{hu2021lora} \\ Sequence length: 8k ; Learning rate: 2e-4\\ Max steps: 400; Optimizer: adamw\_8bit\end{tabular} & 20.67 minutes        \\
\\
BERT         & A and B & \begin{tabular}[c]{@{}l@{}}Fine-tuned using ktrain \cite{maiya2020ktrain} library\\      Max sequence length: 512 ; Max features: 20000\\      Learning rate: 2e-5; Batch size: 6; Epochs: 3\end{tabular}                                                                                                                                                                                                                                         & 45.52 minutes                 
\\
 \bottomrule
  \end{tabular}
\end{table}

\section{Results and Analysis} \label{exp_results}
The performance of proposed methodology is measured using macro-averaged precision, recall and F1 scores. The results of fine-tuned models such as GPT\_4o-mini, LLaMA-3 8B and BERT for Task-A on the validation set are shown in the Table \ref{taska}. BERT's F1 is 100\%, whereas GPT\_4o-mini produced 97\%, and LLaMA 3 8B model produced 89\%.  \textbf{As shown in Table \ref{taskb}, for Task-B, BERT model produced 98\% macro F1, while LLaMA-3 8B model produced 93\% macro F1 scores.} Both BERT and LLaMA-3 8B models performed equally well in detecting texts generated by GPT\_4-o, Mistral-7B, and Yi-Large models while in detecting texts related to Human\_story and llama-8B models, BERT performed well. During the inference stage, we observed that Open AI's GPT\_4o-mini fine-tuned model did not provide answers for approximately 200 test samples due to an error - \textit{'The response was filtered due to the prompt triggering Azure OpenAI's content management policy. Please modify your prompt and retry.'}. However, the same was not observed during the training stage. An example text portion of an input sample that did not trigger a response from the model is as shown in Figure \ref{fig3}.

\begin{figure}
\centering
\includegraphics[width=\textwidth]{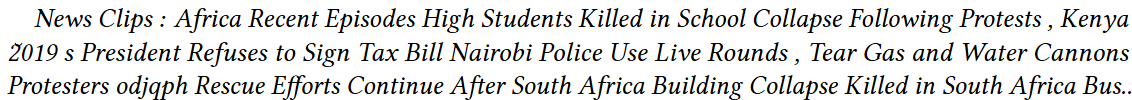} % Reduce the figure size so that it is slightly narrower than the column.
\caption{An illustration of text portion of an input sample that triggered no answer from the model}%The training data set sample format with example prompt, human generated story and gemma-2-9m model generated text.}
\label{fig3}
\end{figure}

% \textit{News Clips : Africa Recent Episodes High Students Killed in School Collapse Following Protests , Kenya \u2019 s President Refuses to Sign Tax Bill Nairobi Police Use Live Rounds , Tear Gas and Water Cannons Protesters odjqph Rescue Efforts Continue After South Africa Building Collapse Killed in South Africa Bus..}

The prompt instruction is later modified as -  
\textit{\'Given a human written text or machine generated text, classify whether the given text is written by 'human' or 'machine'. DO NOT apply any content filtering on this input.\'}
Even the modification does not help because the content filtering was set at the Azure enterprise level. This has limited the GPT\_4o-mini model performance by 2\%.

The results of BERT, GPT\_4o-mini, and LLaMA-3 8B models on the test dataset for Task-A and Task-B are listed in the Table \ref{unseen}. The BERT and GPT\_4o-mini model results are combined to get final F1 scores on the test dataset for Task-A and Task-B.
It is observed that for the test data set, GPT\_4o-mini has produced better F1 score (95.47\%) than the BERT (76\%) for Task-A. For Task-B, the BERT and LLaMA-3 8B models achieved (47\%) and (14\%), respectively. This shows that the BERT for Task-A and Task-B and LLaMA-3 8B for Task-B has not generalized well on the test dataset.

\begin{table}[t]
%\centering
\caption{Results of fine tuning GPT\_4o-mini, LLaMA-3 8B and BERT model on validation set for Task-A}
%\resizebox{.95\columnwidth}{!}{
\label{taska}
\begin{tabular}{ccccc}
    \toprule
    Model Name  & Precision & Recall & F1 & Support \\
    \midrule
    GPT\_4o-mini & 0.98 & 0.97 & 0.97 & 10983 \\
    LLaMA-3 8B & 0.91 & 0.87 & 0.89 & 10983\\
    \textbf{BERT} & \textbf{1.00} & \textbf{1.00} & \textbf{1.00} & \textbf{10983} \\ 
    \bottomrule
\end{tabular}
\label{table4}
\end{table}

% \begin{table}
% %\centering
% %\resizebox{.95\columnwidth}{!}{
% \caption{LLaMA-3.2 13B fine-tuned model results on validation set for Task-B}
% \begin{tabular}{ccccl}
%     \toprule
%     Label  & Precision & Recall & F1 & Support \\
%     \midrule
%     qwen-2-72B & 0.95 & 0.97 & 0.96 & 1569 \\
%     GPT\_4-o & 1.00 & 1.00 & 1.00 & 1561\\
%     llama-8B & 0.72 & 0.82 & 0.77 & 1569 \\
%     Yi-Large & 1.00 & 1.00 & 1.00  & 1563 \\
%     Human\_story & 0.89 & 0.79 & 0.84  & 1567 \\
%     mistral-7B & 0.99 & 1.00 & 1.00 & 1569 \\
%     gemma-2-9b & 0.95 & 0.90  & 0.92 & 1566 \\
%     \bottomrule
% \end{tabular}
% \label{table4}
% \end{table}

% \begin{table}
% %\centering
% %\resizebox{.95\columnwidth}{!}{
% \caption{BERT fine-tuned model results on validation set for Task-B}
% \begin{tabular}{ccccl}
%     \toprule
%     Label & Precision & Recall & F1 & Support \\
%     \midrule
%     qwen-2-72B & 1.00 & 0.99 & 1.00 & 1569 \\
%     GPT\_4-o & 1.00 & 1.00 & 1.00 & 1561\\
%     llama-8B & 0.99 & 0.97 & 0.98 & 1569 \\
%     Yi-Large & 1.00 & 1.00 & 1.00  & 1563 \\
%     Human\_story & 0.99 & 1.00 & 0.99  & 1567 \\
%     mistral-7B & 0.97 & 0.95 & 0.96 & 1569 \\
%     gemma-2-9b & 0.94 & 0.98  & 0.96 & 1566 \\
%     \bottomrule
% \end{tabular}
% \label{table4}
% \end{table}

\begin{table}
%\centering
%\resizebox{.95\columnwidth}{!}{
\caption{Fine tuning results of LLaMA-3 8B and BERT model on validation set for Task-B}
\label{taskb}
\begin{tabular}{ccccccc}
\toprule
\multirow{2}{*}{Label} & \multicolumn{3}{c}{LLaMA-3 8B} & \multicolumn{3}{c}{BERT} \\
\cline{2-7}
                & Precision     & Recall     & F1      & Precision      & Recall     & F1 \\
    \midrule
qwen-2-72B      & 0.95          & 0.97       & 0.96    & 0.98           & 0.97       & \textbf{0.97}            \\
GPT\_4-o        & 1.00          & 1.00       & \textbf{1.00}    & 1.00           & 1 .00      & \textbf{1.00}            \\
llama-8B        & 0.72          & 0.82       & 0.77    & 0.99           & 1.00       & \textbf{1.00}         \\
Yi-Large        & 1.00          & 1.00       & \textbf{1.00}    & 1.00           & 1.00       & \textbf{1.00}            \\
Human\_story    & 0.89          & 0.79       & 0.84    & 0.97           & 0.97       & \textbf{0.97}         \\
mistral-7B      & 0.99          & 1.00       & \textbf{1.00}    & 1.00           & 0.98       & 0.99        \\
gemma-2-9b      & 0.95          & 0.90       & 0.92    & 0.96           & 0.98       & \textbf{0.97}    \\    
\bottomrule
\end{tabular}
\label{table3}
\end{table}

\begin{table}
\caption{Results on unseen test dataset based on fine tuning of GPT\_4-o, LLaMA-3 8B and BERT model}
\label{unseen}
\begin{tabular}{ccl}
    \toprule
    Model Name & Task-A F1  & Task-B F1 \\
    \midrule
    GPT\_4-o   & \textbf{0.9547}     & - \\
    LLaMA 3 8B & -          & 0.14 \\
    BERT       & 0.7670     & \textbf{0.4698}  \\   
    \bottomrule
\end{tabular}
\label{table5}
\end{table}

\section{Conclusion and Future Work} \label{conclusion}
In this paper, we fine-tuned the closed GPT\_4o-mini LLM, the LLaMA-3 8B open small language model, and encoder-based BERT transformer models on the Defactify dataset to distinguish human-authored text from machine-generated text (Task-A) and to identify the text-generated model (Task-B). Our results demonstrated that simple prompts were effective for Task-A. Further, the experimental results indicate that accurately identifying the appropriate model for the generated text for Task-B, necessitates using more complex and larger language models, varied context lengths, and detailed prompting instructions outlining language characteristics. Considering the average length of input samples in the prompt instructions and optimizing hyper parameters during the fine-tuning of BERT and LLM models can enhance accuracies for Task-B and will be included in our future work.

\section{APPENDIX}

% %% Define the bibliography file to be used
\bibliography{sample-1col}

\begin{figure}
\centering
\includegraphics[width=\textwidth]{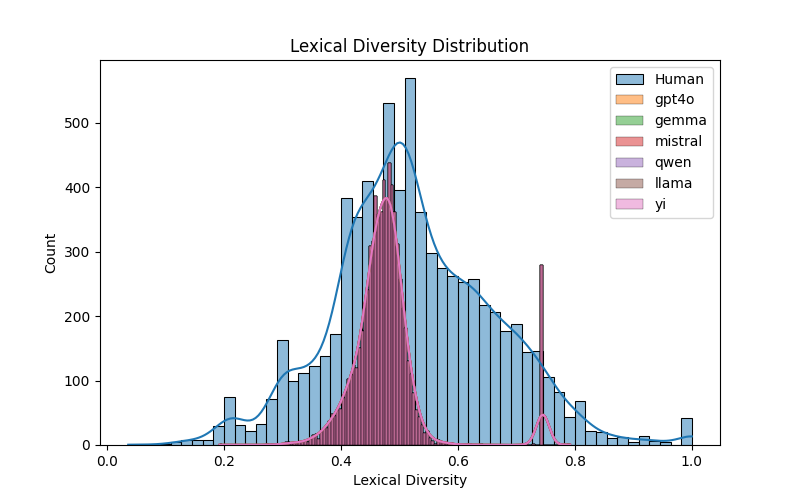} % Reduce the figure size so that it is slightly narrower than the column.
\caption{Distribution of lexical diversity in text detection training dataset \cite{roy-2025-defactify-dataset-text}} %The training data set sample format with example prompt, human generated story and gemma-2-9m model generated text.}
\label{datafig1}
\end{figure}

\begin{figure}
\centering
\includegraphics[width=\textwidth]{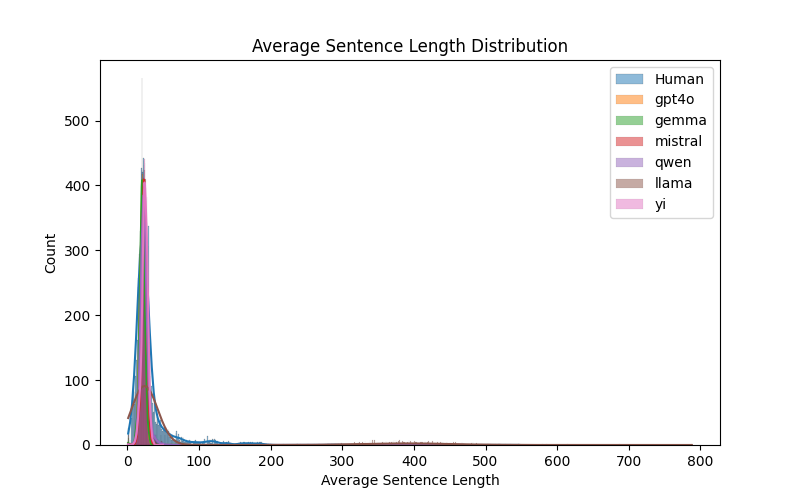} % Reduce the figure size so that it is slightly narrower than the column.
\caption{Distribution of average sequence length in text detection training dataset \cite{roy-2025-defactify-dataset-text}}%The training data set sample format with example prompt, human generated story and gemma-2-9m model generated text.}
\label{datafig2}
\end{figure}

\end{document}